\documentclass[lettersize,journal]{IEEEtran}
\usepackage{xcolor}
\definecolor{iccvblue}{rgb}{0.21,0.49,0.74}
\usepackage{amsmath,amsfonts,amsthm}
\usepackage{algorithmic}
\usepackage{algorithm}
\usepackage{array}
\usepackage{textcomp}
\usepackage{stfloats}
\usepackage{url}
\usepackage{verbatim}
\usepackage{graphicx}
\usepackage{cite}
\usepackage[pagebackref,breaklinks,colorlinks,allcolors=iccvblue]{hyperref}
\usepackage{epigraph} 
\usepackage{makecell}
\usepackage{booktabs}
\usepackage{multirow}
\usepackage{multicol}
\usepackage[dvipsnames]{xcolor}
\usepackage{colortbl}
\usepackage{caption}
\usepackage{subcaption}

\newcommand{\et}[2]{${#1}^{\pm{#2}}$}
\newcommand{\etb}[2]{$\mathbf{{#1}}^{\pm{#2}}$}

\newcommand{\etbb}[2]{$\textcolor{blue}{{#1}}^{\pm{#2}}$}
\newcommand{\etss}[2]{$\textcolor{red}{{#1}}^{\pm{#2}}$}
\newcommand{\atb}[1]{\textcolor{blue}{#1}}
\newcommand{\ats}[1]{\textcolor{red}{#1}}

\def\eg{\emph{e.g.}}
\def\ie{\emph{i.e.}}

\begin{document}

\title{MoSa: $\mathrm{Mo}$tion Generation with $\mathrm{S}$calable $\mathrm{A}$utoregressive Modeling}

\author{
Mengyuan Liu, 
Sheng Yan, 
Yong Wang, 
Yingjie Li,
Gui-Bin Bian,
Hong Liu

\thanks{Mengyuan Liu and Hong Liu are with the State Key Laboratory of General Artificial Intelligence, Peking University,  Shenzhen Graduate School. (e-mail: nkliuyifang@gmail.com; 
hongliu@pku.edu.cn)} 

\thanks{Sheng Yan and Yong Wang are with Chongqing University of Technology. (e-mail: eanson023@gmail.com; ywang@cqut.edu.cn)} 

\thanks{Yingjie Li is with Tencent Technology Co., Ltd. (e-mail: wallaceyjli@tencent.com)} 

\thanks{Gui-Bin Bian is with the State Key Laboratory of Multimodal Artificial Intelligence Systems, Institute of Automation, Chinese Academy of Sciences (e-mail: guibin.bian@ia.ac.cn)}

\thanks{This work was supported by National Natural Science Foundation of China (No. 62473007), Natural Science Foundation of Guangdong Province (No. 2024A1515012089), Shenzhen Innovation in Science and Technology Foundation for The Excellent Youth Scholars (No. RCYX20231211090248064).}} 

\markboth{Submission to IEEE Transactions on Visualization and Computer Graphics}%
{Shell \MakeLowercase{\textit{et al.}}: A Sample Article Using IEEEtran.cls for IEEE Journals}

\IEEEpubid{0000--0000/00\$00.00~\copyright~2025 IEEE}

\maketitle

\begin{abstract}
We introduce MoSa, a novel hierarchical motion generation framework for text-driven 3D human motion generation that enhances the Vector Quantization-guided Generative Transformers (VQ-GT) paradigm through a coarse-to-fine scalable generation process. In MoSa, we propose a Multi-scale Token Preservation Strategy (MTPS) integrated into a hierarchical residual vector quantization variational autoencoder (RQ-VAE). MTPS employs interpolation at each hierarchical quantization to effectively retain coarse-to-fine multi-scale tokens. With this, the generative transformer supports Scalable Autoregressive (SAR) modeling, which predicts scale tokens, unlike traditional methods that predict only one token at each step. Consequently, MoSa requires only 10 inference steps, matching the number of RQ-VAE quantization layers. To address potential reconstruction degradation from frequent interpolation, we propose CAQ-VAE, a lightweight yet expressive convolution-attention hybrid VQ-VAE. CAQ-VAE enhances residual block design and incorporates attention mechanisms to better capture global dependencies. Extensive experiments show that MoSa achieves state-of-the-art generation quality and efficiency, outperforming prior methods in both fidelity and speed. On the Motion-X dataset, MoSa achieves an FID of 0.06 (versus  MoMask’s 0.20) while reducing inference time by 27\%. Moreover, MoSa generalizes well to downstream tasks such as motion editing, requiring no additional fine-tuning. The code is available at \url{https://mosa-web.github.io/MoSa-web}
\end{abstract}
\section{Introduction}
\label{sec:intro}

\IEEEPARstart{T}{ext-driven} 3D human motion generation is a novel and significant branch of human analysis~\cite{zhu2023human,guo2020action2motion,martinez2017human,guo2023back,petrovich2023tmr,yan2023cross,wang2024text,yan2024mlp}, which boasts a wide range of commercial applications. Our showcased program\footnote{\url{https://huggingface.co/spaces/MoSa-web/MoSa}} shows an intuitive example: game designers can perform character modeling without relying on complex motion capture equipment. 
This approach greatly reduces labour and resource costs.

\begin{figure}[t]
\centering
\includegraphics[width=1.0\columnwidth, trim=0mm 0mm 0mm 0mm, clip]{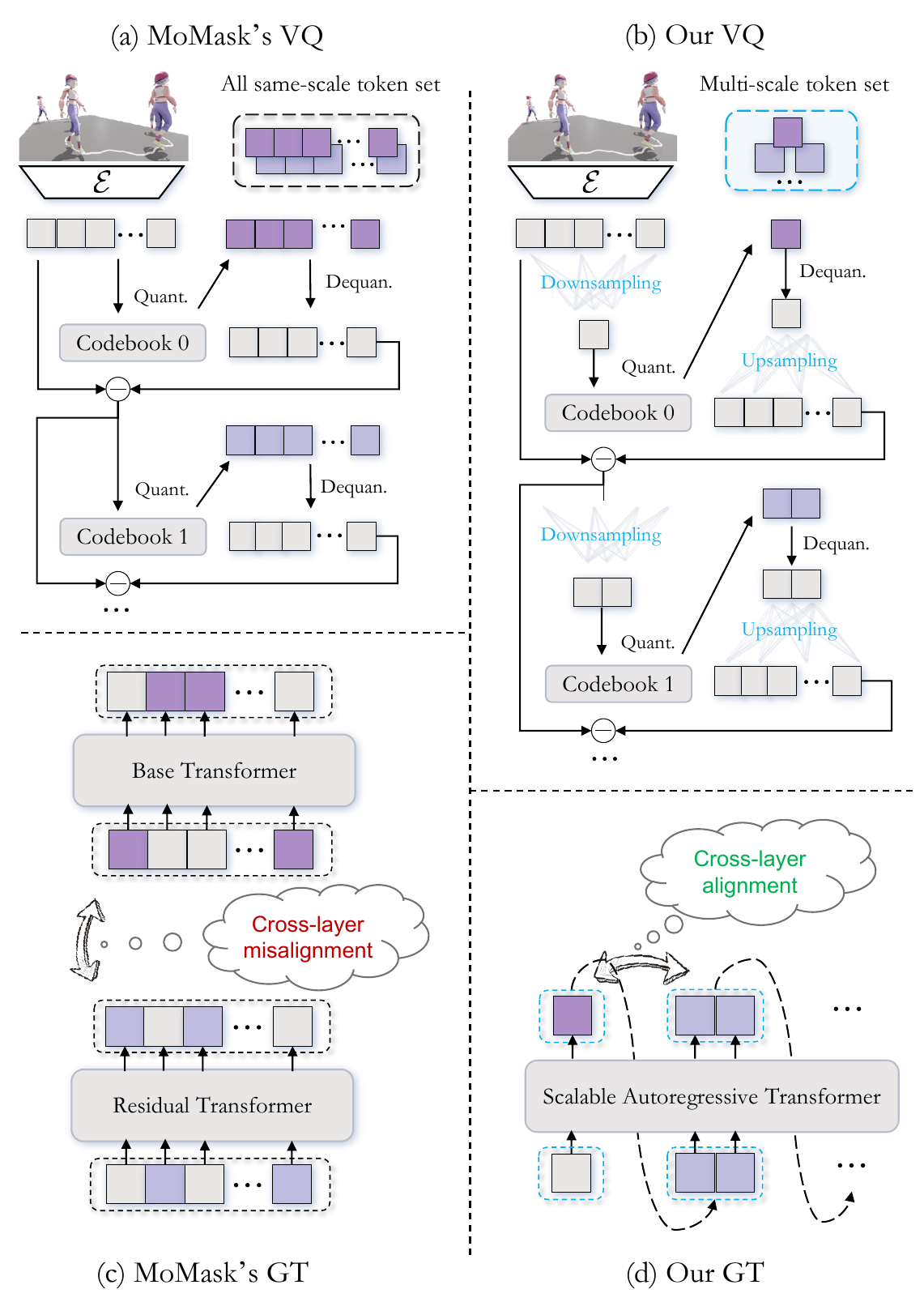}
\caption{Comparison between state-of-the-art method MoMask~\cite{guo2024momask} and our MoSa in the VQ-GT processes: (a) MoMask’s VQ. (b) Our VQ maintains a multi-scale token set via our proposed MTPS, which employs interpolation (downsample/upsampling) at each hierarchical quantization. (c) MoMask’s GT process relies on two independent transformers, leading to cross-layer misalignment. (d) Our GT process with a scalable autoregressive transformer shows cross-layer alignment.}
\label{fig:fig1}
\end{figure}

Consequently, motion generation has attracted considerable research interest~\cite{ahuja2019language2pose,ghosh2021synthesis,guo2022tm2t,jiang2023motiongpt,kong2023priority,zhang2023generating,guo2024momask,tevet2022motionclip,petrovich2022temos,guo2022generating,chen2023executing,meng2024rethinking}. Earlier works like TEMOS~\cite{petrovich2022temos} and MotionCLIP~\cite{tevet2022motionclip} aimed to fit the distribution between semantics and motion. Following the success of diffusion models~\cite{ho2020denoising}, numerous studies~\cite{zhang2023remodiffuse,yuan2023physdiff,zhang2022motiondiffuse,karunratanakul2023guided,zhou2024emdm} shifted towards diffusion-based motion generation, such as the representative MLD~\cite{chen2023executing}, which conducts diffusion within the latent space. In parallel, another paradigm, combining motion vector quantization~\cite{van2017neural, esser2021taming, williams2020hierarchical} and generative transformers~\cite{vaswani2017attention, devlin2019bert, radford2019language} (called VQ-GT) in a two-stage framework~\cite{guo2020action2motion, jiang2023motiongpt,guo2022tm2t,kong2023priority}, has achieved competitive performance. \eg, T2M-GPT~\cite{zhang2023generating} quantizes motion into specialized discrete tokens and utilizes a transformer to generate continuous human motion. 
\IEEEpubidadjcol
However, the VQ process inherently introduces approximation errors, which led the latest state-of-the-art method, MoMask~\cite{guo2024momask}, to improve generation precision by introducing a hierarchical residual vector quantization variational autoencoder (RQ-VAE)~\cite{lee2022autoregressive} to preserve fine-grained details (see Fig.~\ref{fig:fig1} (a) MoMask's VQ). During the GT process, MoMask separates the hierarchical tokens into base (first-layer) and residual parts, and employs two independent masked transformers~\cite{chang2022maskgit} to model them. While this approach offers stronger representation power compared to single-layer VQ-VAE, each layer of tokens is generated independently, leading explicit cross-layer misalignment (see Fig.~\ref{fig:fig1} (c) MoMask's GT). As a result, if the residual transformer fails to capture the structural context of the input tokens, it may lead to incoherent detail refinement.

To better exploit intermediate representations, we propose a novel framework, MoSa, that enhances the VQ-GT paradigm through a coarse-to-fine scalable generation process. In MoSa, we first introduce a Multi-scale Token Preservation Strategy (MTPS) integrated into the RQ-VAE. MTPS leverages interpolation at each hierarchical quantization level to effectively retain multi-scale tokens from coarse to fine. With this, our autoregressive transformer is capable of jointly modeling all intermediate representations during the GT phase, thereby mitigating the aforementioned cross-layer misalignment and enhancing the global modeling capacity of the transformer. 

Precisely, in VQ, as illustrated in Fig.\ref{fig:fig1}(b), unlike previous methods\cite{guo2024momask,pinyoanuntapong2025bamm} that directly preserve all same-scale intermediate tokens, our Multi-scale Token Preservation Strategy 2(MTPS) maintains a hierarchical token set across multiple scales. Starting from the first layer of the residual quantizer, motion sequences \textit{downsampling} into coarse-scale representations. These are then quantized, with \textit{upsampling} to align with the original motion length. As the hierarchy deepens, finer scales are progressively introduced, with tokens from each scale retained until reaching the final granularity (\ie, 49 tokens corresponding to motion latent length). MTPS enables us to extend the classic autoregressive (AR) modeling paradigm into a Scalable-Autoregressive (SAR) modeling. As shown in Fig.~\ref{fig:fig1}(d), during training, the transformer jointly learns from the entire multi-scale token set by scanning from coarse to fine scales. At inference time, instead of generating one token at a time as in conventional AR, the model predicts multiple tokens of the next scale in parallel. Consequently, the total number of inference steps is reduced to the number of RQ-VAE quantization layers, which is set to 10 in our experiments, allowing the generation process to be completed in just 10 steps.

Although MTPS brings significant improvements in inference speed, its frequent interpolation operations (\ie, downsampling and upsampling) introduce detail distortions during motion reconstruction. To address this critical issue, we further optimize the encoder-decoder architecture. Specifically, we propose CAQ-VAE, a lightweight yet expressive convolution-attention hybrid VQ-VAE. CAQ-VAE enhances the design of residual blocks and incorporates attention mechanisms to better capture global dependencies. Notably, the model size of CAQ-VAE remains comparable to prior methods. Experimental results show that MoSa achieves approximately a 27\% improvement in inference speed while maintaining state-of-the-art generation quality. For example, on the latest and largest Motion-X dataset, MoSa achieves an FID of 0.06 (vs. 0.20 by MoMask), demonstrating superior overall performance.

In addition, we explore the extensibility of MoSa. We demonstrate that MoSa can also be applied to motion editing tasks such as motion inpainting and outpainting without any additional fine-tuning. The model achieves strong qualitative results in these settings as well. In summary, our main contributions are as follows:
\begin{itemize}
\item We propose MoSa, which introduces a Multi-scale Token Preservation Strategy (MTPS) to retain motion tokens across different scales. This strategy enables Scalable-Autoregressive (SAR) modeling, which jointly models all intermediate representations and generates motions in a coarse-to-fine manner.
\item We propose CAQ-VAE, a lightweight yet expressive convolution-attention hybrid VQ-VAE that mitigates detail distortion during motion reconstruction.
\item We explore the extensibility of MoSa and show that it generalizes well to motion editing tasks without requiring additional fine-tuning.
\end{itemize}

\begin{figure*}[htbp]
\centering
\includegraphics[width=1.0\textwidth]{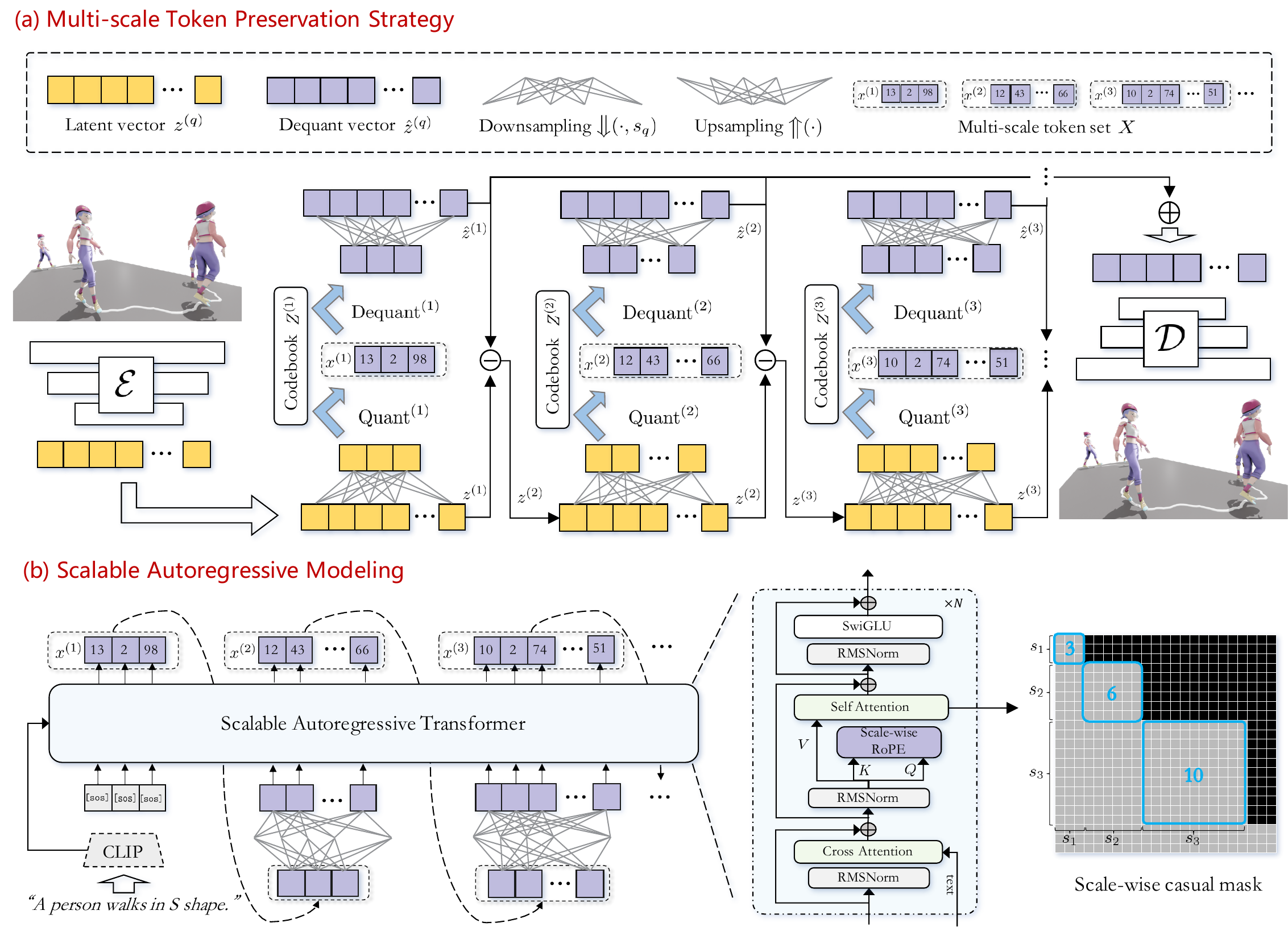}
\vspace{-15pt}
\caption{\textbf{Our MoSa framework overview.}  
(a) Multi-scale Token Preservation Strategy (MTPS) integrated into a hierarchical RQ-VAE. MTPS employs interpolation (Downsampling/Upsampling operation) at each hierarchical quantization to effectively retain coarse-to-fine multi-scale token set $X$. The scales follow a predefined scheduler \( S = (s_{1}, s_{2}, s_{3}, \dots, s_{Q}) \), where \( s_{q} \leq T \), representing a coarse-to-fine hierarchy. The illustration shows an example with $(s_{1}=3, s_{2}=6, s_{3}=10)$. 
(b) The multi-scale token set supervise Scalable Autoregressive (SAR) modeling. Given an input \( (\texttt{[sos]}, x^{(1)}, x^{(2)}, \dots, x^{(Q-1)}) \), the SAR predicts \( (x^{(1)}, x^{(2)}, \dots, x^{(Q)}) \), where multiple tokens within each scale are predicted in parallel. During training, a scale-wise attention mask ensures that each \( x^{(q)} \) can only attend to \( x^{\leq (q)} \). Notably, the \( x^{(q)} \) contains \( s_q \) tokens, while \( x^{(q-1)} \) has only \( s_{(q-1)} \) tokens. Before feeding \( x^{(q-1)} \) into the Transformer, the \( x^{(q-1)} \) will be Up-Downsampling to match \( s_q \).  As illustrated, the input representation of $x^{(2)}$ is derived from up-downsampling $x^{(1)}$, and $x^{(3)}$ from $x^{(2)}$.}
\label{fig:model}
\end{figure*}

\section{Related Work}
\label{sec:relatedwork}

\noindent\textbf{Text-driven motion generation.} Text possesses strong semantic expressiveness, enabling precise descriptions of various actions, speeds, and directions, making it a key modality for human motion generation~\cite{petrovich2021action,guo2020action2motion,ghosh2021synthesis, tang2018dance, le2023music, ginosar2019learning, alexanderson2023listen, zhu2023taming}. Early approaches, \eg, Text2Action~\cite{ahn2018text2action}, leveraged GANs to generate diverse motions from natural language descriptions. JL2P~\cite{ahuja2019language2pose} employed a GRU-based encoder-decoder framework to map text to corresponding human motions. To tackle zero-shot generation, MotionCLIP~\cite{tevet2022motionclip} aligned motion latent spaces with the text and image embeddings of the pre-trained CLIP model, significantly improving zero-shot generalization. TEMOS~\cite{petrovich2022temos} further optimized the joint multimodal latent space via a VAE.  
Inspired by the success of text-to-image generation, diffusion models~\cite{ho2020denoising} and VQ-VAE~\cite{van2017neural} have been widely adopted for text-to-motion generation. The former introduces a forward diffusion process that gradually corrupts data, training a network to recover motions via reverse diffusion~\cite{zhang2023remodiffuse,yuan2023physdiff,zhang2022motiondiffuse,karunratanakul2023guided,zhou2024emdm, kim2023flame}. The latter, exemplified by TM2T~\cite{guo2022tm2t} and T2M-GPT~\cite{zhang2023generating}, discretizes human motions into tokens via VQ-VAE and employs a GPT-like transformer for autoregressive generation~\cite{jiang2023motiongpt,zhang2024motiongpt,pinyoanuntapong2024mmm}. MoMask further refines this approach by introducing hierarchical quantization and leveraging BERT-style masked modeling~\cite{devlin2018bert, chang2022maskgit} to train both base and residual transformers, achieving state-of-the-art performance.  In this work, we follow the VQ-VAE paradigm and demonstrate that competitive performance can be achieved with just a two-stage training process.


\noindent\textbf{Autoregressive models.} In image synthesis~\cite{hartwig2025survey, su2022drawinginstyles, chen2025human, huang2025creativesynth, lu2024facemug, zeng2022aggregated}, autoregressive models have leveraged insights from NLP by using VQ-VAE to quantize images into tokens and employing transformers to predict them~\cite{esser2021taming,yu2021vector,yu2022scaling} sequentially. However, this token-by-token approach does not align well with the autoregressive assumption for images with inherently complex spatial structures. VAR~\cite{tian2024visual}, building upon~\cite{lee2022autoregressive}, innovatively reformulates token prediction as scale prediction. This scalable modeling strategy predicts all tokens within a specific scale at once, helping to maintain internal consistency in image generation. Further developments include~\cite{li2024controlvar}, which introduced a controllable framework, and~\cite{zhang2024var, ma2024star}, which explored scale-based generation for text-to-image synthesis. xAR~\cite{ren2025beyond} demonstrates that the prediction units do not necessarily have to be scaled—they can also be fixed regions or arbitrary subsamples. Inspired by these advances, we introduce scalable modeling into human motion synthesis. To the best of our knowledge, we leverage the latest scalable modeling techniques for the first time.

\section{Preliminary}
\label{lab:preliminary}

\subsection{Vector Quantization}
\label{lab:vq}
Human motion is inherently represented as a continuous signal.  
To apply autoregressive modeling to motion (see Sec.~\ref{lab:autoregressive}), we need to convert it into discrete tokens.  This is typically achieved using a vector quantized autoencoder (VQ-VAE), \eg, T2M-GPT~\cite{zhang2023generating}, which converts motion latent features \( z \in \mathbb{R}^{T \times C} \) into discrete tokens \( x \in [V]^{T} \): 
\begin{align}  
    z = \mathcal{E}(m), \quad  
    x = \mathrm{Quant}(z)  
\end{align}  
where \( m \) denotes the original motion, \( \mathcal{E}(\cdot) \) is the encoder, and \( \mathrm{Quant}(\cdot) \) is the quantization function.  
The quantization needs a learnable codebook \( Z \in \mathbb{R}^{V \times C} \) containing \( V \) vectors, which aiming to maps each latent vector \( z_{(t)} \) to its nearest code index \( x_{(t)} \) based on Euclidean distance:  
\begin{align}
    x_{(t)} = \left( \text{argmin}_{v \in [V]} \| Z_v - z_{(t)} \|_2 \right) \in [V]
\end{align}  
Given the discrete tokens $x$, the corresponding codebook embeddings can be retrieved through a dequantization function \( \mathrm{Dequant}(Z, \cdot) \), which maps token indices back to code vectors \( \hat{z} \). The decoder \( \mathcal{D}(\cdot) \) then reconstructs motion \( \hat{m} \) from \( \hat{z} \), and the optimization minimizes a compound loss \( \mathcal{L}_{\text{vq}} \):  
\begin{align}  
\begin{aligned}
    \hat{z} &= \mathrm{Dequant}(Z, x), \quad  
    \hat{m} = \mathcal{D}(\hat{z}), \\  
    \mathcal{L}_{\text{vq}} &= \|m - \hat{m} \|_1 + \|sg[z] - \hat{z}\|_2 + \beta \|z - sg[\hat{z}]\|_2  
    \label{eq:vaeloss}  
\end{aligned}
\end{align}  
where \( sg[\cdot] \) denotes the stop-gradient operation, and \( \beta \) is the weight of the embedding constraint. The entire process is optimized using the straight-through gradient estimator~\cite{van2017neural}, and the codebook \( Z \) is updated via exponential moving averages and codebook resets. 
 
\subsection{Classic Autoregressive
Modeling}
\label{lab:autoregressive}
Consider a sequence of discrete tokens \( x = (x_1, x_2, \dots, x_T) \), where \( x_t \in [V] \) drawn from the VQ-VAE aforementioned. 
Classic autoregressive methods assume that the probability of the current observation \( x_t \) depends on its preceding context \( (x_1, x_2, \dots, x_{t-1}) \) and text condition \(c\).  
This unidirectional token dependency allows the likelihood of the sequence \( x \) to be factorized as:  
\begin{small}
\begin{align}  
    p(x_1, x_2, \dots, x_T \mid c) = \prod_{t=1}^{T} p(x_t \mid x_1, x_2, \dots, x_{t-1}, c) \label{eq:ar}  
\end{align}  
\end{small}

\noindent Training an autoregressive model \( p_\theta \) involves optimizing the conditional probability \( p_\theta(x_t \mid x_1, x_2, \dots, x_{t-1}, c) \) over the dataset. Once trained, \( p_\theta \) can be used to generate new sequences.

\section{Our MoSa}

Although the Preliminary is effective, the VQ process inevitably introduces approximation errors. To address this, some methods generate a \textit{set} of same-scale discrete tokens by quantizing the residuals~\cite{guo2024momask,pinyoanuntapong2025bamm}, known as Residual VQ-VAE (RQ-VAE). The core idea is to reduce overall quantization error through iterative residual quantization progressively. This process requires \( Q \) quantizers instead of a single one:  
\begin{align}  
\begin{aligned}
    x^{(q)} &= \mathrm{Quant}^{(q)}(z^{(q)}),  \quad 
    \hat{z}^{(q)} = \mathrm{Dequant}(Z^{(q)}, x^{(q)})  \label{eq:quant}  
\end{aligned}
\end{align} 
and $z^{(q+1)} = z^{(q)} - \hat{z}^{(q)}$, for \( q = 1, \dots, Q \). After quantization, the final approximation \( \hat{z} \) is obtained as the sum of all dequantizations \(\sum_{q=1}^Q \hat{z}^{(q)}\).
To train RQ-VAE, the optimization objective should integrate the constraints of all quantizers:  
\begin{align}  
\begin{aligned}  
    \mathcal{L}_{\text{commit}} &=\sum_{q=1}^Q \left( \|sg[z^{(q)}] - \hat{z}^{(q)}\|_2 + \beta \|z^{(q)} - sg[\hat{z}^{(q)}]\|_2 \right)  , \\
    \mathcal{L}_{\text{rvq}} &= \|m - \hat{m} \|_1 + \mathcal{L}_{\text{commit}}\label{eq:rqvaeloss}  
\end{aligned}
\end{align}  
\noindent This process creates a same-scale discrete token set \( (x^{(1)}, \dots, x^{(Q)}), x^{(q)} \in [V]^{T} \), which provides supervision for training generative models. \eg, the MoMask uses a base transformer to 
 model the first layer tokens \( x^{(1)} \) and a residual transformer for the rest \( x^{(2):(Q)} \).

In the rest of this section, we first discuss our Multi-scale Token
Preservation Strategy (Sec.~\ref{subsec:mtps}) and the Scable Autoregressive modeling (Sec.~\ref{subsec:sar}). Then, we present the detailed Convolution-Attention hybrid VQ-VAE architecture in Sec.~\ref{subsec:caq-ave}. Finally, we discuss the motion editing tasks applications in Sec. \ref{subsec:downstream}.

\subsection{Multi-scale Token Preservation Strategy}  
\label{subsec:mtps}
 
Unlike the previous approach of saving all same-scale intermediate layer tokens, our MTPS maintains a multi-scale token set: 
\begin{small}
\begin{align} 
    X\hspace{-2pt} = \hspace{-2pt}\left\{\underbrace{(x^{1}_1, \dots, x^{1}_{s_{1}})}_{x{^{(1)}}}, \underbrace{(x^{2}_1, x^{2}_2, \dots, x^{2}_{s_{2}})}_{x{^{(2)}}}, \dots, \underbrace{(x^{Q}_1, x^{Q}_2, \dots, x^{Q}_{s_{Q}})}_{x{^{(Q)}}} \right\}  
\end{align}  
\end{small}
This set consists of \(Q\) scales: \( S = (s_{1}, s_{2}, \dots, s_{Q}) \), where \( s_{q} \le T \). \eg, \( S = (3, 6, \dots, 49) \) represents a predefined schedule that moves from a coarse to a fine scale. The final scale \(s_{Q}\) matches the motion latent length \( T \) representing the fine scale. 

As illustrated in Fig.~\ref{fig:model}(a), a Downsampling operation \({\big\Downarrow}(\cdot, s_{q})\) is performed to reduce the latent vector $z^{(q)}$ from fine scale \( s_{Q} \) to scale \( s_{q} \) before each residual quantization step:
\begin{small}
\begin{align}  
    x^{(q)} = \,\mathrm{Quant}^{(q)}({\big\Downarrow}(z^{(q)}, s_{q})), \, 
    \hat{z}^{(q)} = \,{\big\Uparrow}(\mathrm{Dequant}^{(q)}(Z^{(q)}, x^{(q)}))
\end{align}
\end{small}
This design aims to obtain compact tokens $x^{(q})$ at the specific scale $s_{q}$. Following this, an Upsampling operation \({\big\Uparrow}(\cdot)\) is then applied after dequantization to recover the approximated value \(\hat{z}^{(q)}\). In contrast to the common RQ-VAE (Eq.~\ref{eq:quant}), the incorporation of interpolation operations enables the generation of compact tokens \( x^{(q)} \in [V]^{s_{q}} \) at specific scales, rather than producing all the same-scale tokens.

Within \(Q\) times quantization, the scale \(s_{q}\) is progressively increased while storing the tokens until the scale reaches a fine level \(s_{Q}\). This allows for the maintenance of a multi-scale token set \(x\), enabling Scalable Autoregressive modeling mentioned in the next section (see Sec.~\ref{subsec:sar}). The overall objective $\mathcal{L}_{\text{rvq}}$ remains unchanged.  

\subsection{Scalable Autoregressive Modeling}  
\label{subsec:sar}
By maintaining the multi-scale token sets, our autoregressive transformer is capable of jointly modeling all intermediate representations in the GT phase, thereby mitigating the cross-layer misalignment and enhancing the global modeling capacity of the transformer. We reformulate the autoregressive modeling (Sec.~\ref{lab:autoregressive}) into Scalable Autoregressive (SAR) modeling as shown in Fig.~\ref{fig:model}(b). Here, the autoregressive unit is scale tokens rather than a single one. The SAR likelihood is defined as: 

\vspace{-10pt}
\begin{small}
\begin{align}  
    p(x^{(1)}, x^{(2)}, \dots, x^{(Q)} \mid c) = \prod_{q=1}^{Q} p(x^{(q)} \mid x^{(1)}, x^{(2)}, \dots, x^{(q-1)}, c) \label{eq:scalable}  
\end{align}  
\end{small}

\noindent where \( x^{(q)} = (x^{(q)}_1, x^{(q)}_2, \dots, x^{(q)}_{s_q}) \) represents the specific scale tokens predicted at the \( q \)-th autoregressive step. Notably, SAR generates multiple tokens simultaneously at each step, distinguishing it from traditional autoregressive methods (Eq. \ref{eq:ar}) that predict only a single one. The sequence \( (x^{(1)}, x^{(2)}, \dots, x^{(q-1)}) \) and the condition \( c \) serve as the ``prefix" for \( x^{(q)} \). In the \( q \)-th step, the distribution of all \( s_q \) tokens in \( x^{(q)} \) is generated in parallel, conditioned on its prefix and corresponding positional embeddings.

Note that \( x^{(q)} \) contains \( s_q \) tokens, while \( x^{(q-1)} \) has only \( s_{(q-1)} \) tokens. Before feeding \( x^{(q-1)} \) into the Transformer to generate the distribution of \( x^{(q)} \), the \( x^{(q-1)} \) will be up-downsampling to match \( s_q \). Furthermore, during training, MoSa employs a scale-wise causal attention mask, ensuring that each \( x^{(q)} \) can only attend to its prefix as presented in Fig.~\ref{fig:model}(b)-right.   

\noindent\textbf{KV caching allowed again.} During inference, our model retains the autoregressive property. The KV cache technology is reintroduced, and no mask is needed. The inference steps correspond to the multi-scale set size \( Q \) (\ie, the RQ-VAE quantization layers), avoiding token-by-token decoding. 

\begin{figure}[t]
\centering
\includegraphics[width=0.95\columnwidth]{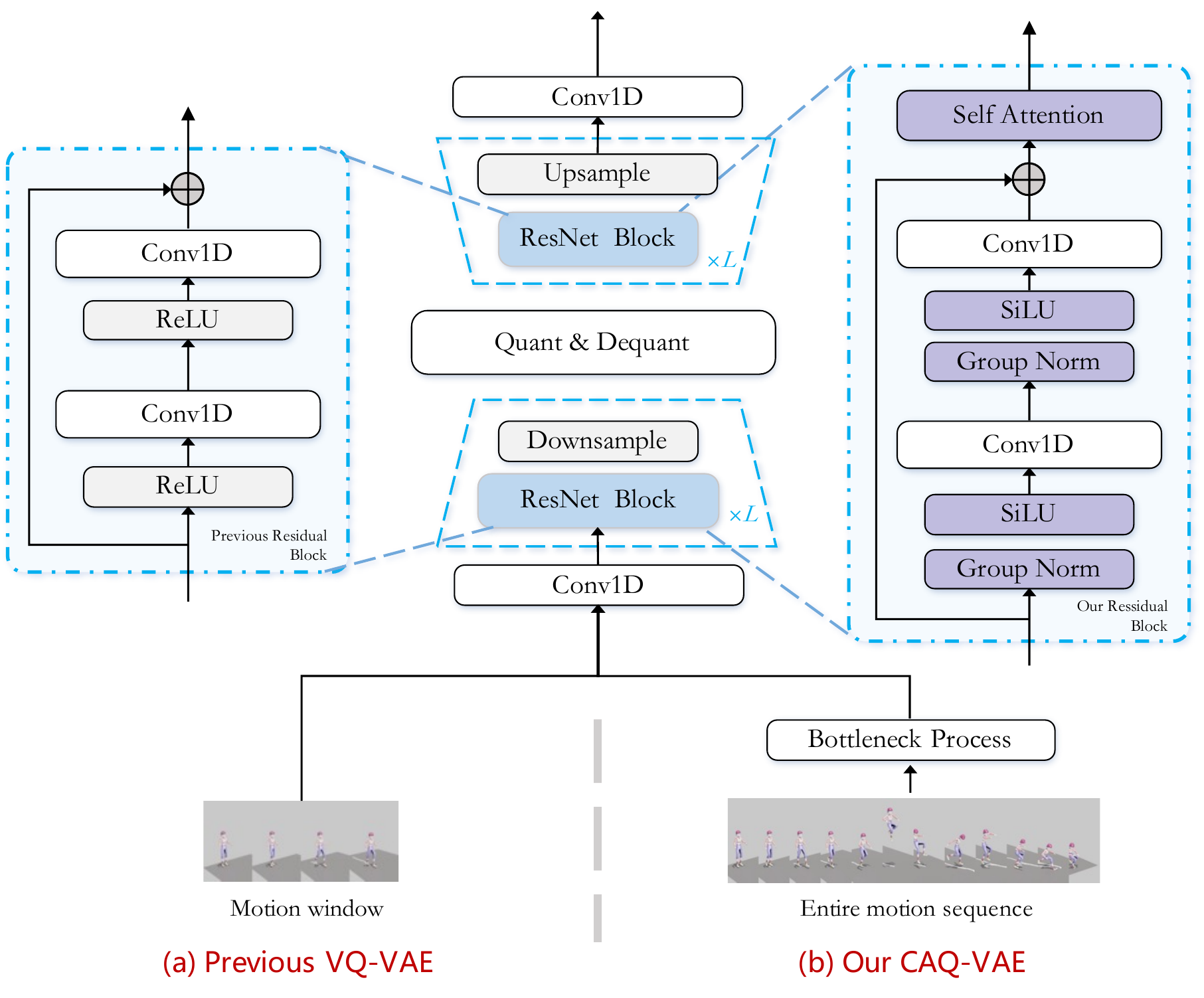}
\caption{\textbf{Previous VQ-VAE compared to our CAQ-VAE.} Our CAQ-VAE uses residual blocks with GroupNorm and SiLU, along with a self-attention layer to capture global dependencies.}
\label{fig:network}
\end{figure}

\noindent\textbf{Transformer.} Our architecture closely aligns with LLaMA, incorporating RMSNorm~\cite{zhang2019root} and SwiGLU activations~\cite{shazeer2020glu}. Besides, we employ a target perturbation strategy from machine translation to perturb the input sequence \(x\), mitigating the training-inference discrepancy. This strategy is also used in T2M-GPT. Additionally, word-level text embeddings interact with motion via cross-attention (see Fig.~\ref{fig:network}(b)) to address the issue of neglecting textual information when the transformer optimizes tokens across all scales. Finally, the standard cross-entropy loss is used, with increased weight on the final scale's optimization, to enhance the quality of the final generated output. 

\noindent\textbf{Scale-wise RoPE.} Rotary position embedding (RoPE)~\cite{su2024roformer} encodes absolute and relative positions via complex rotations. To adapt RoPE to our multi-scale structure, we redefine token positions relative to their scale. For a scale $s_q$, the original position $m$ is normalized as $\frac{m}{s_q} \times s_{Q}$.

\subsection{Convolution-Attention hybrid VQ-VAE}  
\label{subsec:caq-ave}

Although MTPS brings significant improvements in inference speed, its frequent interpolation operations (\ie, downsampling and upsampling) introduce detail distortions during motion reconstruction. To address this critical issue, we further optimize the VQ-VAE encoder-decoder architecture with a lightweight yet expressive convolution-attention hybrid VQ-VAE (CAQ-VAE). 

\noindent\textbf{Architecture.} The prior VQ-VAE divides the motion sequence into 64-frame windows and reconstructs it through convolutions to accelerate training~\cite{guo2022generating, guo2024momask}. However, this strategy conflicts with MTPS, which requires perceiving the entire sequence at multiple scales. Therefore, CAQ-VAE takes the entire motion sequence as input, which naturally motivates the employment of attention to model global dependencies. Additionally, the residual blocks in the prior VQ-VAE lack normalization, which may limit expressiveness. To address this, CAQ-VAE adopts GroupNorm for stable feature distribution and replaces ReLU with SiLU to enhance nonlinearity. Lastly, we introduce a Bottleneck Process that expands and then compresses the intermediate channel dimensions, improving modeling capacity without increasing the parameter count.

\noindent\textbf{Recovery net}. Unlike VAR in the vision domain, which inserts shared residual blocks after dequantization to compensate for detail distortion, we find that a simple non-shared two-layer convolutional network with ReLU achieves a good balance between reconstruction quality and generation diversity.

\noindent\textbf{\(\ell_2\)-norm.} We apply \(\ell_2\) normalization~\cite{yu2021vector} during codebook quantization, transforming the Euclidean distance into cosine similarity, which enhances codebook usage. This approach is commonly used in recent vision reconstruction tasks~\cite{yu2021vector,tian2024visual,yu2024randomized}.

\noindent\textbf{Codebook scaling.} Unlike previous VQ-VAE and VAR methods that use a shared codebook across all quantization layers, our CAQ-VAE adopts non-shared codebooks to enhance representational capacity. Additionally, the codebook sizes $V$ are linearly increased across layers. We find that using smaller codebooks in the earlier layers lowers the difficulty of subsequent SAR prediction.

\subsection{Motion Editing}

Benefiting from our SAR modeling, motion generation at each scale can attend to both intra-scale and preceding scale context. Leveraging this design, we further explore a compelling application of our model—Motion Editing—which requires no additional training.

Motion Editing encompasses a variety of sub-tasks, including motion inpainting, outpainting, prefix filling, suffix filling, and free-form motion completion. Specifically, we introduce an $\mathrm{MASK}_{\text{edit}}$ to specify the regions for generation, while the remaining tokens are replaced with ground-truth tokens obtained from our CAQ-VAE. During inference, the $\mathrm{MASK}_{\text{edit}}$ is interpolated to ensure coherent and consistent editing across scales.

Notably, the free-form motion completion sub-task operates without language conditioning and is guided purely by classifier-free guidance. The visualization of the results will be shown in our experiment section.

\label{subsec:downstream}

\section{Experiments}

\begin{table*}[]
\centering
\renewcommand{\arraystretch}{1.4}
\caption{\textbf{Quantitative evaluation on the 
HumanML3D and Motion-X test set.} $\pm$ indicates a 95\% confidence interval. \textcolor{blue}{Blue} and \textcolor{red}{Red} indicate the best and the second best result. `$\dagger$' denotes our reimplementation. The results of MoMask are slightly inconsistent with those reported in the paper (shown in \textcolor{gray}{gray}). The relevant issue has been discussed in \href{https://github.com/EricGuo5513/momask-codes/issues/27}{https://github.com/EricGuo5513/momask-codes/issues/27} as well as in~\cite{wang2025spatial}.}
\label{tab:main_results}
\resizebox{\textwidth}{!}{%
\begin{tabular}{llcccccc|ccc}
\toprule
\multirow{2}{*}{Datasets} & \multirow{2}{*}{Methods} & \multicolumn{3}{c}{R Precision$\uparrow$} & \multirow{2}{*}{FID$\downarrow$} & \multirow{2}{*}{MultiModal Dist$\downarrow$} & \multirow{2}{*}{MultiModality$\uparrow$} & \multirow{2}{*}{Stage$\downarrow$} & \multirow{2}{*}{Step $\downarrow$} & \multirow{2}{*}{AIT$\downarrow$} \\ \cline{3-5}
 &  & Top 1 & Top 2 & Top 3 &  &  &  &  & &  \\ \midrule

\multirow{7}{*}{\makecell[c]{Human\\ML3D}} & \textbf{Real motions} & \et{0.511}{.003} & \et{0.703}{.003} & \et{0.797}{.002} & \et{0.002}{.000} & \et{2.974}{.008} & - & - & - & - \\ \cline{2-11} 
 & TEMOS~\cite{petrovich2022temos} & \et{0.424}{.002} & \et{0.612}{.002} & \et{0.722}{.002} & \et{3.734}{.028} & \et{3.703}{.008} & \et{0.368}{.018} & \atb{1} & \atb{1} & \atb{0.016} \\
  & MotionDiffuse~\cite{zhang2022motiondiffuse} & \et{0.491}{.001} & \et{0.681}{.001} & \et{0.782}{.001} & \et{0.630}{.001} & \et{3.113}{.001} & \et{1.553}{.042} & \atb{1} & 1,000 & 4.086 \\
 & T2M-GPT~\cite{zhang2023generating} & \et{0.492}{.003} & \et{0.679}{.002} & \et{0.775}{.002} & \et{0.141}{.005} & \et{3.121}{.009} & \etss{1.831}{.048} & \ats{2} & 49 &  0.127  \\
 & MLD~\cite{chen2023executing} & \et{0.481}{.003} & \et{0.673}{.003} & \et{0.772}{.002} & \et{0.473}{.013} & \et{3.196}{.010} & \etbb{2.413}{.079} & \ats{2} & 50 & 0.094 \\
 & MoMask$^\dagger$~\cite{guo2024momask} & \etss{0.504}{.003} & \etss{0.699}{.003} & \etss{0.795}{.002} & \etss{0.124}{.006} & \etss{3.062}{.010} & \et{1.327}{.044}& 3 & 15 & 0.062 \\ 
 & \textcolor{gray}{MoMask~\cite{guo2024momask}} & \textcolor{gray}{\et{0.521}{.002}} & \textcolor{gray}{\et{0.713}{.002}} & \textcolor{gray}{\et{0.807}{.002}} & \textcolor{gray}{\et{0.045}{.002}} & \textcolor{gray}{\et{2.958}{.008}} & \textcolor{gray}{\et{1.241}{.040}} & \textcolor{gray}{3} & \textcolor{gray}{15} & \textcolor{gray}{0.062} \\ 
 & \textbf{MoSa (Ours)} & \etbb{0.530}{.003} & \etbb{0.725}{.002} & \etbb{0.821}{.003} & \etbb{0.085
}{.003} & \etbb{2.836}{.009} & \et{1.763}{.059} & \ats{2} & \ats{10} & \ats{0.045} \\ \midrule 
 
 \multirow{7}{*}{\makecell[c]{Motion-\\X}} & \textbf{Real motions}$^\dagger$ & \et{0.480}{.002} & \et{0.699}{.002} & \et{0.812}{.002} & \et{0.001}{.000} & \et{2.682}{.003} &- & - & - & - \\ \cline{2-11} 
 & TEMOS$^\dagger$~\cite{petrovich2022temos} & \et{0.290}{.001} & \et{0.467}{.002} & \et{0.584}{.002}  & \et{6.448}{.004} & \et{4.923}{.008} & \et{0.435}{.031} & \atb{1} & \atb{1} & \atb{0.016} \\
  & MotionDiffuse$^\dagger$~\cite{zhang2022motiondiffuse} & \et{0.387}{.002} & \et{0.589}{.003} & \et{0.714}{.002} & \et{1.980}{.036} & \et{3.521}{.013} & \et{2.155}{.074} & \atb{1} & 1,000 & 4.086 \\
 & T2M-GPT$^\dagger$~\cite{zhang2023generating} & \et{0.385}{.003} & \et{0.571}{.004} & \et{0.679}{.002} & \et{0.974}{.045} & \et{3.855}{.019} & \etss{2.429}{.122} & \ats{2} & 49 & 0.127 \\
 & MLD$^\dagger$~\cite{chen2023executing} & \et{0.415}{.002} & \et{0.618}{.003} & \et{0.734}{.002} & \et{0.463}{.008} & \et{3.421}{.003} & \etbb{2.597}{.078} & \ats{2} & 50 & 0.094 \\
  & MoMask$^\dagger$~\cite{guo2024momask} & \etss{0.439}{.002} & \etss{0.647}{.002} & \etss{0.760}{.002} & \etss{0.200}{.004} & \etss{3.131}{.009} & \et{1.501}{.075} & 3 & 15 & 0.062 \\
 & \textbf{MoSa (Ours)} & \etbb{ 0.448}{.003} & \etbb{0.657}{.003} & \etbb{0.771}{.002} & \etbb{0.061}{.003} & \etbb{2.982}{.007} & \et{1.754}{.062} & \ats{2} & \ats{10} & \ats{0.045} \\ \bottomrule 
\end{tabular}%
}
\end{table*}

In this section, we present the results of our experiments. We introduce the datasets and evaluation protocol in Sec. \ref{lab:db_and_evalm}. Subsequently, we compare our results with competing methods' results in Sec. \ref{lab:comparision}, followed by related ablation experiments in Sec. \ref{lab:ablation}. Then, we present the coarse-to-fine generation process in Sec. \ref{subsec:generation}. Finally, we show the extension application in the motion editing task in Sec.~\ref{subsec:motion_editing}.

\subsection{Experimental Setup}
\label{lab:db_and_evalm}

\begin{figure}[t]
 \centering
  \begin{subfigure}[b]{0.49\columnwidth}
    \includegraphics[width=\linewidth]{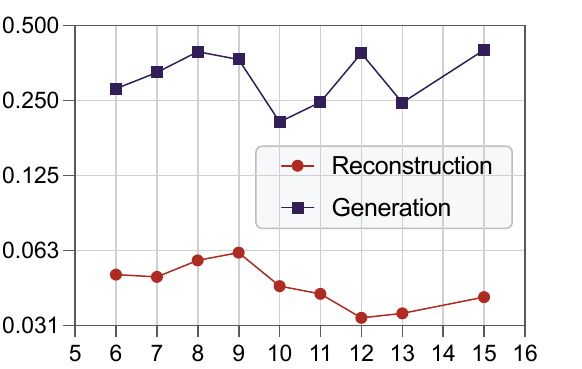}
    \caption{FID$\downarrow$}
    \label{fig:fid}
  \end{subfigure}
  \hfill 
  \begin{subfigure}[b]{0.49\columnwidth}
    \includegraphics[width=\linewidth]{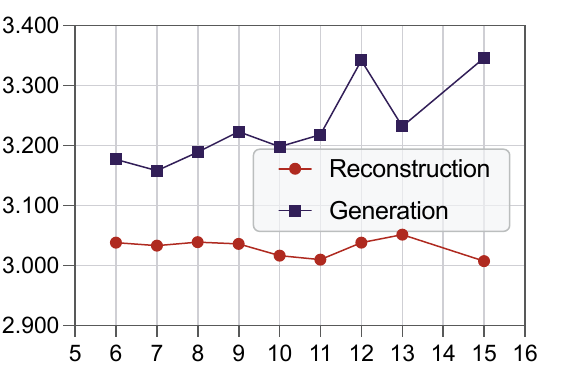}
    \caption{MM-Dist$\downarrow$}
    \label{fig:mmdist}
  \end{subfigure}
  \hfill 
  \begin{subfigure}[b]{0.49\columnwidth}
    \includegraphics[width=\linewidth]{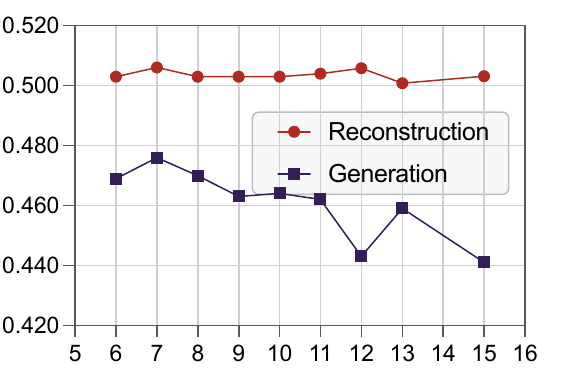}
    \caption{Top-1$\uparrow$}
    \label{fig:top1}
  \end{subfigure}
  \hfill 
  \begin{subfigure}[b]{0.49\columnwidth}
    \includegraphics[width=\linewidth]{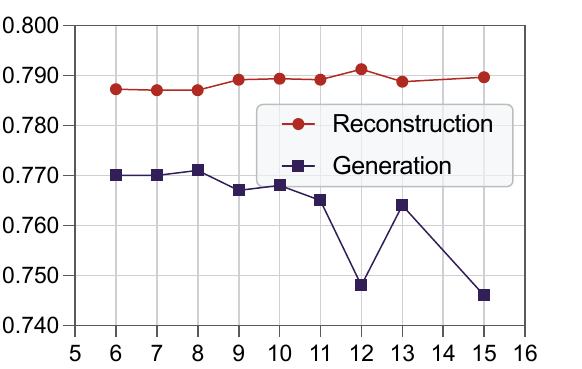}
    \caption{Top-3$\uparrow$}
    \label{fig:top3}
  \end{subfigure}
  \caption{\textbf{Impact of multi-scale token set size on HumanML3D.} Using the MoSa-mini, we trained both the VQ Model (\textit{Reconstruction} task) and the Transformer for text-to-motion synthesis (\textit{Generation} task) on the HumanML3D dataset. The x-axis represents the size of the multi-scale token set $Q$, which also determines the total inference steps (ranging from 6 to 15). The results indicate that $Q=10$ achieves the best overall balance across all metrics.}
  \label{fig:q}
\end{figure}

\begin{figure*}[ht]
\centering
\includegraphics[width=1.0\linewidth, trim=0mm 0mm 5mm 0mm, clip]{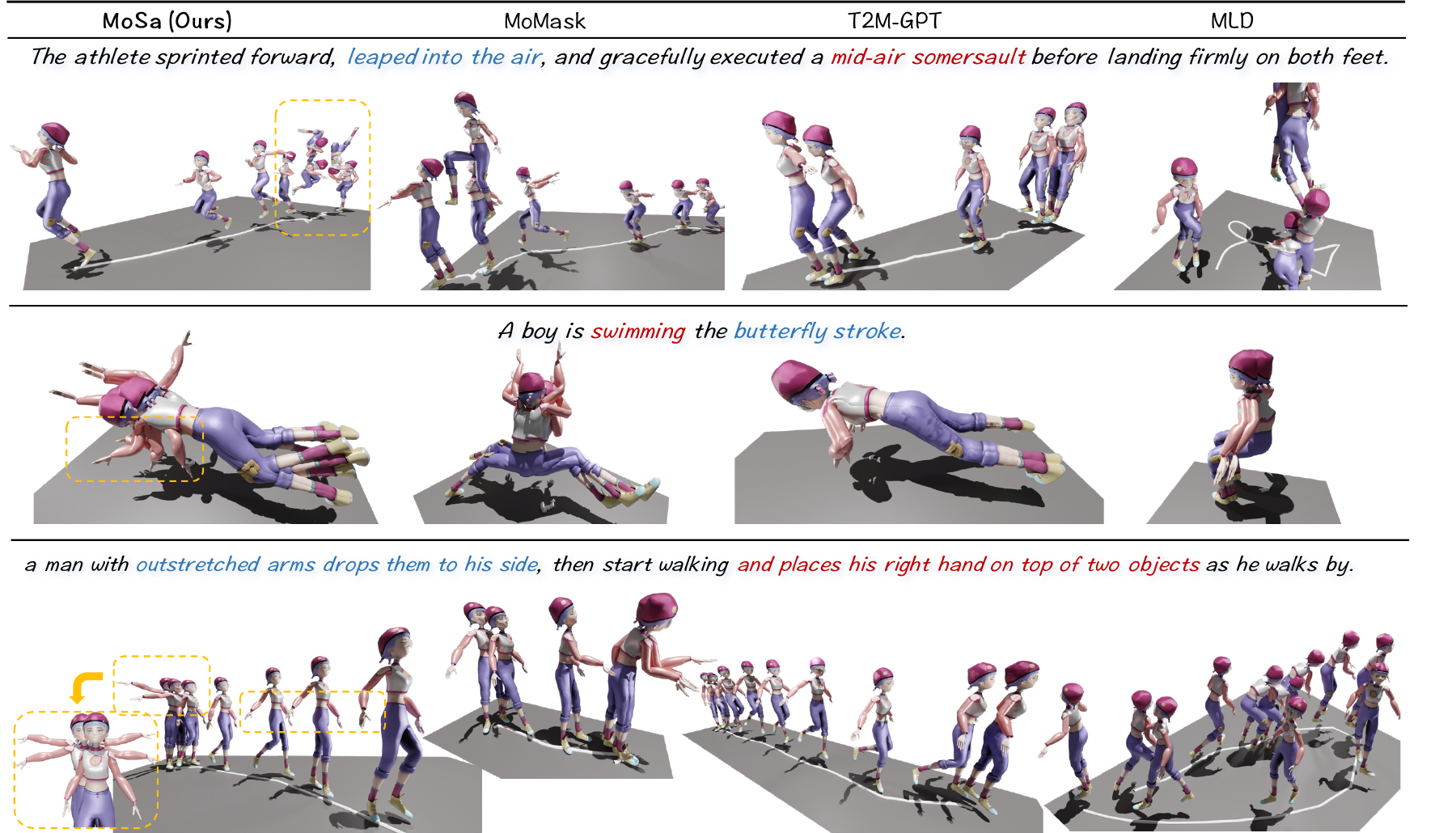}
\caption{\textbf{Qualitative evaluation on Motion-X dataset.} Motions that align with key semantics are highlighted in \textcolor{Goldenrod}{yellow}. For more dynamic visualizations, please refer to the project page.}
\label{fig:vs_results}
\end{figure*}

We conduct experiments on two motion-text datasets: 
HumanML3D~\cite{guo2022generating}, and the latest, larger-scale Motion-X dataset~\cite{lin2023motion}. We follow the most evaluation protocol proposed in \cite{guo2022generating}.  

\textbf{HumanML3D} is a medium-scale dataset that includes 14,616 high-quality motions paired with 44,970 text descriptions, where each motion is described by three different captions. \textbf{Motion-X} is the most recent and largest motion-text dataset, featuring greater diversity. Following the protocol of the first dataset, we filter out motion-text pairs exceeding 200 frames, resulting in 37,751 motion sequences and 61,637 text captions. The datasets are split into training, validation, and test sets with an 80\%, 5\%, and 15\% ratio, respectively.  

To ensure consistency, both datasets are represented using the guo263 format~\cite{guo2022generating}. That is, the whole-body representation in Motion-X is converted. Since most text descriptions primarily focus on body movements, we omit hand and facial features to prevent unnecessary modality discrepancies. Additionally, we train a feature extractor on it to evaluate generation quality. The training code is largely based on~\cite{guo2022generating}.

\noindent\textbf{Implementation details.}  
We use CLIP-ViT-B/32~\cite{radford2021learning} to extract word embedding. For the CVQ-VAE, the VQ requires \(Q=10\) quantizers. For the transformer, we use 8 layers, 6 heads, and a latent dimension of 384. The dimension of the SwiGLU is set to 768. The learning rate is linearly warmed up, reaching 2e-4 after 2,000 iterations. The VQ Model is trained with a batch size of 256, while the transformer is trained with 
64 for HumanML3D and Motion-X. During inference, the classifier-free guidance (CFG)~\cite{ho2022classifier}  scale is set to 4 for both datasets. The CFG scale decays as the scale $s_q$ increases.

\noindent\textbf{Evaluation metrics.} We adopt the following evaluation metrics:  
(1) the \textit{Frechet Inception Distance (FID)}, which assesses the overall action quality by measuring the distributional difference between the high-level features of generated and real actions; 
(2) \textit{R-Precision} and \textit{Multimodal distance}, which are used to measure the semantic consistency between the input text and the generated actions; 
(3) \textit{Multimodality}, which is used to evaluate the diversity of actions generated from the same text.
(4) \textit{Average inference time (AIT)}, which quantifies the model’s inference efficiency.

\subsection{Comparison to state-of-the-art approaches}
\label{lab:comparision}

We compare our MoSa with existing state-of-the-art baseline methods:
(1) TEMOS~\cite{petrovich2022temos}: Utilizes a variational autoencoder (VAE) trained on motion data to generate compatible latent space distribution parameters.
(2) T2M-GPT~\cite{zhang2023generating}: Learns discrete motion representations and employs a GPT-like prediction mechanism using a CLIP prior.
(3) MotionDiffuse~\cite{zhang2022motiondiffuse}: Introduces diffusion models for motion generation.
(4) MLD~\cite{chen2023executing}: Adapts the latent diffusion model to learn motion representations for a VAE.
(5) MoMask~\cite{guo2024momask}: Uses two bidirectional transformers to capture base and residual discrete representations.

\noindent\textbf{Quantitative comparisons.}  
Table \ref{tab:main_results} presents the quantitative results across the two datasets. The AIT represents the average inference time (Seconds), measured on an NVIDIA RTX 4090, averaged over 100 samples.  We also tip the training stage (Stage) and inference steps (Step) for each method. For Motion-X, we reproduce and evaluate the baseline methods using our trained feature extractor. Their training hyperparameters strictly align with the HumanML3D.
Each experiment is evaluated 20 times, and the mean scores are reported with a 95\% confidence interval. 

MoSa demonstrates superior or competitive performance across a broad range of metrics while maintaining high inference efficiency. On the HumanML3D dataset, MoSa achieves the highest Top-1 (0.530), Top-2 (0.725), and Top-3 (0.821) R-Precision scores, surpassing all existing methods, including our reimplemented MoMask. MoSa also achieves the lowest FID (0.085), outperforming MoMask's 0.124, demonstrating clear superiority in motion quality. Furthermore, MoSa maintains the lowest multimodal distance (2.836) compared to MoMask (3.062), demonstrating superior performance in balancing diversity and realism.

On the Motion-X dataset, MoSa continues to lead across all key quality metrics. It achieves the best FID of 0.061, substantially lower than MoMask (0.200) and significantly better than other baselines such as MLD (0.463) and T2M-GPT (0.974). In terms of R-Precision, MoSa reaches Top-1: 0.448, Top-2: 0.657, and Top-3: 0.771, again outperforming all competitors including MoMask (0.439, 0.647, 0.760). MoSa also achieves a lower MultiModal Dist (2.982) compared to MoMask (3.131), demonstrating better semantic alignment. Notably, MoSa exhibits exceptional inference efficiency. With only 2 autoregressive stages and 10 steps, it achieves an average inference time (AIT) of just 0.045s—faster than MoMask (0.062s), offering a 27\% speedup. Unlike diffusion-based models such as MotionDiffuse, which require 1,000 steps and incur an AIT of 4.086s, MoSa offers a practical and scalable solution for real-time applications. Together, these results validate MoSa's advantage in generating high-quality, diverse, and semantically aligned motions with remarkable efficiency.

\noindent\textbf{Qualitative comparisons.}  Fig.~\ref{fig:vs_results} compares MoSa, MLD~\cite{chen2023executing}, T2M-GPT~\cite{zhang2023generating}, and MoMask~\cite{guo2024momask}, with samples generated from the Motion-X checkpoint to highlight differences. MLD and T2M-GPT capture general meaning but struggle with details. \eg, while generating ``A boy is swimming the butterfly stroke," they produce a basic swimming pose without the characteristic wide arm movements. MoMask shows some improvement but still has alignment issues. In the first instance, it fails to depict the expected ``mid-air somersault." Additionally, the action ``outstretched arms drop to his side" is not accurately represented. More dynamic visualizations are available in the website video.

\begin{figure*}[htbp]
\centering
\includegraphics[width=1.0\linewidth, trim=20mm 0mm 0mm 0mm, clip]{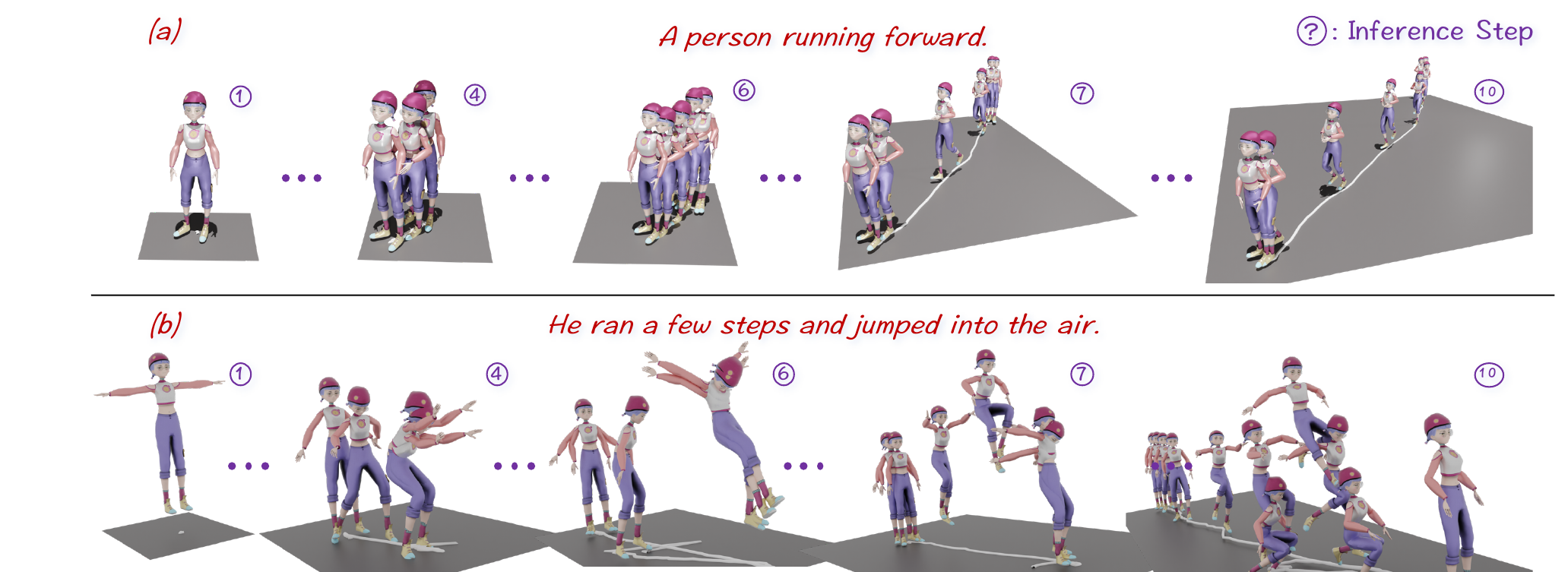}
\caption{\textbf{Visualization of the coarse-to-fine generation process.} Starting at a coarse scale (3 tokens, Step 1) and progressively refined to a fine scale (49 tokens, Step 10). The final representation is achieved through dequantization and upsampling from the multi-scale token set and incremental accumulation into the VQ model for reconstruction.} 
\label{fig:proof}
\end{figure*}

\subsection{Ablation study}
\label{lab:ablation}

\begin{table}[]
\centering
\renewcommand{\arraystretch}{1.2}
\caption{Ablation of our CAQ-VAE model and comparison of previous work on the HumanML3D and Motion-X datasets. The ablation study evaluates the effectiveness of the strategies proposed in ~\ref{subsec:caq-ave}. `$\dagger$' denotes our reimplementation, which are slightly inconsistent with the paper-reported results (shown in \textcolor{gray}{gray}). The relevant issue has been discussed in \href{https://github.com/EricGuo5513/momask-codes/issues/27}{https://github.com/EricGuo5513/momask-codes/issues/27}.
}
\label{tab:ablation_main}
\resizebox{\columnwidth}{!}{%
\begin{tabular}{@{}lcclcc@{}}
\toprule
\multirow{2}{*}{Methods} & \multicolumn{2}{c}{Reconstruction} &  & \multicolumn{2}{c}{Generation} \\ \cmidrule(lr){2-3} \cmidrule(l){5-6} 
 & FID$\downarrow$ & Top 1$\uparrow$  & & FID$\downarrow$ & MM-Dist$\downarrow$ \\ \hline
\rowcolor{Thistle!10}
\multicolumn{6}{c}{\textit{Evaluation on the HumanML3D dataset}} \\ \hline 
T2M-GPT~\cite{guo2022generating} & \et{0.070}{.001} & \et{0.501}{.002}  &  & \et{0.141}{.005} & \et{3.121}{.009} \\
MoMask$^\dagger$~\cite{guo2024momask} & \et{0.032}{.000} & \etb{0.507}{.003} &   & \et{0.124}{.006} & \et{3.062}{.010} \\
\textcolor{gray}{MoMask~\cite{guo2024momask}} & \textcolor{gray}{\etb{0.019}{.001}} & \textcolor{gray}{\etb{0.509}{.002}} &   & \textcolor{gray}{\etb{0.051}{.002}} & \textcolor{gray}{\et{2.957}{.008}} \\
\textbf{MoSa (Our CAQ-VAE)} & \etb{0.030}{.000} & \et{0.507}{.004}  &  & \etb{0.085}{.003} & \etb{2.836}{.009} \\ \,\, \textit{w/o} CA hybrid & \et{0.055}{.002} & \et{0.486}{.003} &   & \et{0.150}{.004} & \et{3.011}{.009} \\ \,\, \textit{w/o} Bottleneck Process & \et{0.032}{.002} & \et{0.506}{.004} &  & \et{0.093}{.003}  & \et{2.849}{.009} \\
\,\, \textit{w/o} Recovery net & \et{0.035}{.002} & \et{0.504}{.004} &  & \et{0.229}{.006}  & \et{3.042}{.009} \\
\,\, \textit{w/o} $\ell_2$-norm & \et{0.042}{.002} & \et{0.503}{.004} &   & \et{0.124}{.006} & \et{2.881}{.009} \\
\,\, \textit{w/o} Codebook scaling & \et{0.040}{.002} & \et{0.504}{.004} &   & \et{0.118}{.006} & \et{2.874}{.009} \\ \hline
\rowcolor{Thistle!10}
\multicolumn{6}{c}{\textit{Evaluation on the Motion-X dataset}} \\ \hline
T2M-GPT~\cite{guo2022generating} & \et{0.170}{.002} & \et{0.425}{.003} & & \et{0.974}{.045} & \et{3.855}{.019} \\
MoMask~\cite{guo2024momask} & \et{0.063}{.001} & \et{0.450}{.002} &  & \et{0.200}{.004} & \et{3.131}{.009} \\
\textbf{MoSa (Our CAQ-VAE)} & \etb{0.027}{.001} & \etb{0.455}{.002}  &  & \etb{0.061}{.003} & \etb{2.982}{.007} \\ \bottomrule
\end{tabular}%
}
\end{table}

\begin{table}[]
\centering
\renewcommand{\arraystretch}{1.2}
\caption{We evaluate the impact of text fusion methods and position encoding (PE) strategies, including RoPE~\cite{su2024roformer} and our proposed Scale-wise RoPE.  The result was evaluated using the HumanML3D test set.}
\label{tab:transformer}
\begin{tabular}{@{}lccc@{}}
\toprule
Module & FID$\downarrow$ & Top 1$\uparrow$ & MM-Dist$\downarrow$ \\ \hline
\rowcolor{Thistle!10}
\multicolumn{4}{c}{\textit{The impact of text fusion method}} \\ \hline
\texttt{[sos]} & \et{0.107}{.004} & \et{0.464}{.003}  & \et{3.237}{.010} \\
AdaLN & \et{0.093}{.004} & \et{0.522}{.003}  & \et{2.954}{.009} \\
Cross attention & \etb{0.085}{.003} & \etb{0.530}{.003}  & \etb{2.836}{.009} \\ \hline
\rowcolor{Thistle!10}
\multicolumn{4}{c}{\textit{The impact of position encoding}} \\ \hline
w/o PE & \et{0.085}{.003} & \et{0.497}{.003} & \et{2.979}{.008} \\
Absolute PE & \et{0.104}{.002} & \et{0.508}{.002} &  \et{2.963}{.008} \\
RoPE & \et{0.086}{.004} & \et{0.518}{.002} & \et{2.900}{.008} \\
Scale-wise RoPE  & \et{0.085}{.003} & \etb{0.530}{.003}  & \etb{2.836}{.009} \\ \bottomrule
\end{tabular}%
\end{table}

\begin{figure}[t]
\centering
\includegraphics[width=\linewidth]{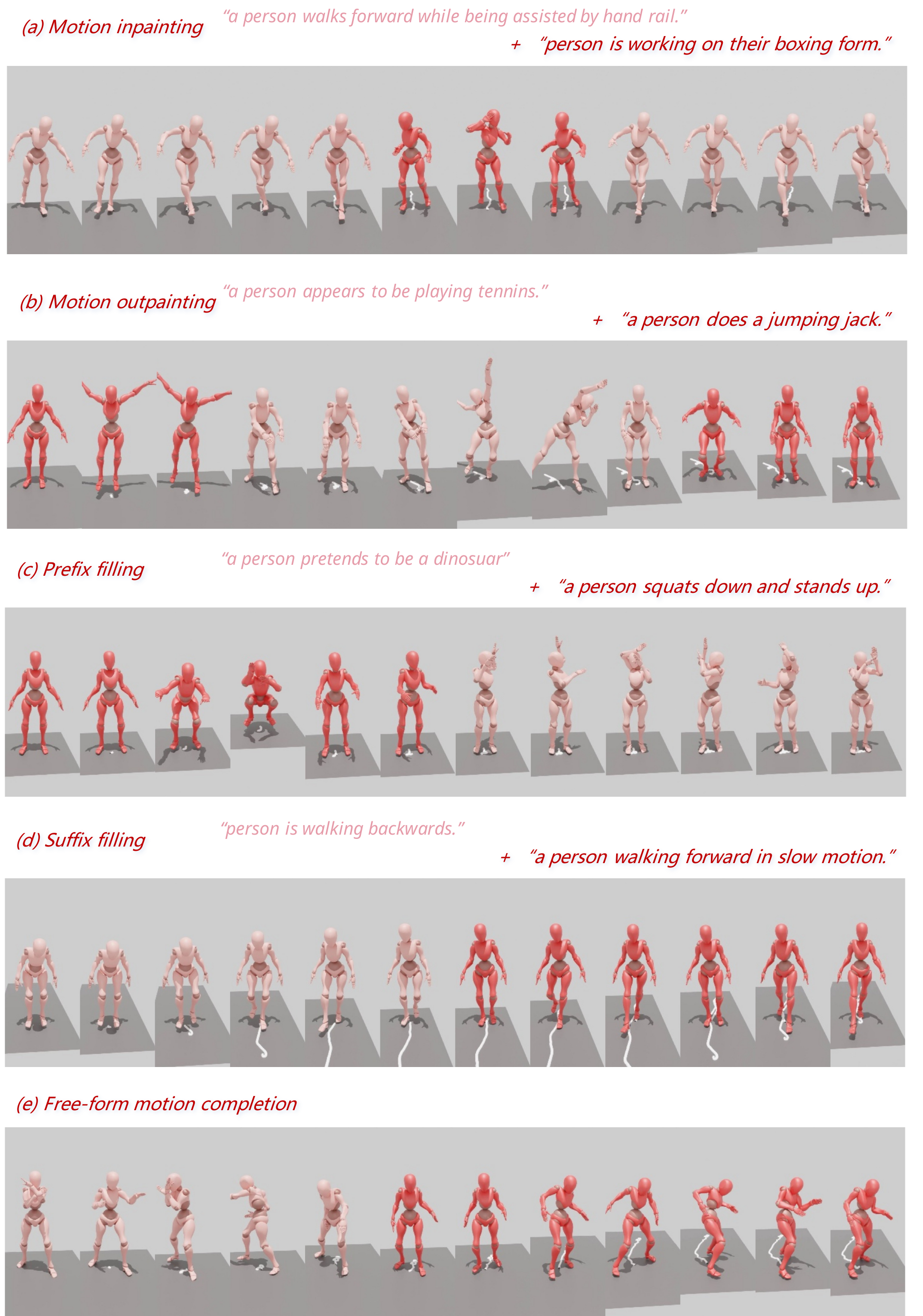}
\caption{\textbf{Visualization of the motion editing.} Motion Editing encompasses a variety of sub-tasks, including motion inpainting, outpainting, prefix filling, suffix filling, and free-form motion completion. The input motion clips are highlighted in \textcolor{pink}{pink}, and the generated motions are depicted in \textcolor{red}{red}. More results on motion editing are available on our project page.}
\label{fig:motion_editing}
\end{figure}

\noindent\textbf{Number of scales $Q$.} We examine the impact of multi-scale set size \( Q \) (\ie, total inference steps) on model performance with the HumanML3D dataset. For efficiency, we use MoSa-mini, a smaller version of MoSa with half the parameters, as our base model. Fig.~\ref{fig:q} shows how different \( Q \) values affect the VQ model (\textit{Reconstruction}) and the Transformer (\textit{Generation}). Increasing \( Q \) improves reconstruction quality, indicated by a lower FID score. However, in generation tasks, too large a \( Q \) results in performance decline; for instance, when \( Q = 15 \), Top-1 precision drops to 44\%. 
This suggests that a larger \( Q \) increases token count, complicating transformer modeling and increasing inference errors. Balancing generation and reconstruction, we select \( Q = 10 \) as the optimal configuration. The predefined scheduler is $S=(3,6,10,15,20,25,30,36,42,49)$. Subsequent experiments are conducted using the full MoSa model.

\noindent\textbf{Ablation on CAQ-VAE.}
Table~\ref{tab:ablation_main} reports the ablation results of our proposed CAQ-VAE on HumanML3D and Motion-X, demonstrating the impact of each component. Removing the convolution-attention hybrid leads to a significant performance drop in both reconstruction (FID: 0.030 → 0.055) and generation (FID: 0.085 → 0.150), highlighting the importance of combining local convolutions with global attention for expressive sequence modeling. Excluding the bottleneck process slightly increases the generation FID from 0.085 to 0.093, indicating its effectiveness in enhancing feature modeling. Our CAQ-VAE encoder-decoder contains only 18.84M parameters, which is 0.6M fewer than the VQ-VAE modules used in T2M-GPT and MoMask. The recovery network, though lightweight, notably reduces generation FID from 0.229 to 0.085, validating its role in compensating detail loss post-quantization. Most notably, removing the $\ell_2$-norm during quantization not only degrades generation quality (FID: 0.085 → 0.124), but also significantly reduces codebook utilization from 99.5\% to 88.9\%, confirming its necessity for encouraging diverse code usage via cosine similarity. Finally, using a shared codebook across layers (i.e., removing codebook scaling) increases generation FID from 0.085 to 0.118, suggesting that scale-wise codebooks help balance quantization difficulty and support the SAR predictor. Together, these results confirm that each design choice in CAQ-VAE contributes to the overall high-fidelity reconstruction and diverse generation performance.

\noindent\textbf{Alation on transformer.} 
Table~\ref{tab:transformer} presents the impact of different design choices in our transformer. First, using cross-attention for text fusion significantly improves performance over the baseline \texttt{[sos]}. This improvement may stem from our transformer’s requirement to model all scale tokens in parallel—236 tokens, which far exceeds T2M-GPT's 49 tokens. Simply adding sentence-level features to the \texttt{[sos]} position risks losing crucial information. Second, Scale-wise RoPE consistently outperforms other position encoding methods, demonstrating its effectiveness in multi-scale token prediction.

\subsection{Analysis in Generation}
\label{subsec:generation}

\begin{figure}[t]
 \centering
  \begin{subfigure}[b]{0.49\columnwidth}
    \includegraphics[width=\linewidth]{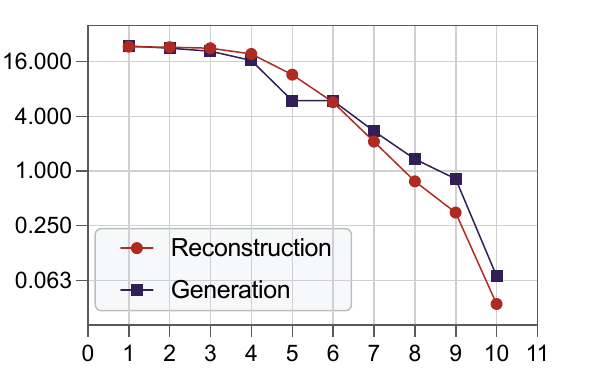}
    \caption{FID$\downarrow$}
    \label{fig:fid-2}
  \end{subfigure}
  \hfill 
  \begin{subfigure}[b]{0.49\columnwidth}
    \includegraphics[width=\linewidth]{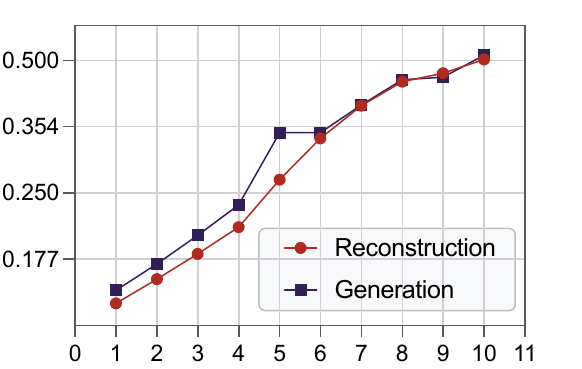}
    \caption{Top-1$\uparrow$}
    \label{fig:top1-2}
  \end{subfigure}
  \caption{\textbf{Step-wise cumulative performance on HumanML3D.} From inference steps 1 to 10, the metrics show a progressive improvement, indicating MoSa's coarse-to-fine characteristics.}
  \label{fig:infer}
\end{figure}

To validate the effectiveness of MoSa's coarse-to-fine generation, we visualize the generation process in Fig.~\ref{fig:proof}. At the initial stage, the generated motion exhibits key poses but lacks proper limb coordination. As the generation progresses, the poses become increasingly natural, with more refined details. For example, in (a) step 7, the walking pose shows a raised leg, which illustrates the gradual refinement of the motion. This visualization effectively demonstrates how MoSa transitions from coarse, low-detail motions to smoother, more realistic poses, validating the success of its coarse-to-fine motion generation approach. In parallel, Fig.~\ref{fig:infer} provides step-wise quantitative evidence: from step 1 to 10, the generation FID improves dramatically from 23.92 to 0.085, while Top-1 accuracy rises from 0.15 to 0.530, confirming that MoSa's inference strategy progressively enhances both fidelity and semantic alignment throughout the generation process.

\subsection{Application: Motion Editing}
\label{subsec:motion_editing}

Benefiting from our SAR modeling, motion generation at each scale can attend to both intra-scale and preceding scale context. Leveraging this design, we further explore a compelling application of our model—Motion Editing—which requires no additional training. Motion Editing encompasses a variety of sub-tasks, including motion inpainting, outpainting, prefix filling, suffix filling, and free-form motion completion. As shown in Fig.~\ref{fig:motion_editing}, we mask 50\% of the motion sequence and use the remaining 50\% for editing. MoSa demonstrates strong scalability and produces smooth transitions at the edited boundaries. Notably, free-form motion completion—which operates without any textual guidance—still generates diverse and high-quality motion sequences. More results on motion editing are available on our project page\footnote{\url{https://mosa-web.github.io/MoSa-web}}.

\section{Conclusion}
\label{sec:conclusion}
We introduce MoSa, a new framework for text-driven 3D human motion generation that improves both the quality of generated motion and the efficiency of inference. By utilizing a refined hierarchical structure and a scalable autoregressive transformer, MoSa generates motion in a coarse-to-fine manner, preserving multi-scale token representations and maintaining consistency between encoding and generation. Our experiments show a reduction in inference time while maintaining competitive generative quality.



\bibliographystyle{IEEEtran}
\bibliography{bare_jrnl_new_sample4}

\end{document}